\documentclass[12pt]{iopart}
\usepackage{iopams}

\expandafter\let\csname equation*\endcsname\relax
\expandafter\let\csname endequation*\endcsname\relax

\usepackage{amsmath}
\usepackage{amssymb}
\usepackage{amsfonts}
\usepackage{mathtools}
\usepackage{graphicx}
\usepackage{textcomp}
\usepackage[utf8]{inputenc}
\usepackage{stackengine}
\usepackage{bm}

\usepackage[margin=1in]{geometry}
\usepackage{todonotes, placeins}
\usepackage{hyperref, xcolor}

\usepackage{subfigure}
\usepackage{bm}
\usepackage{float}
\restylefloat{table}
\usepackage{xcolor,soul}
\usepackage{algorithm}
\usepackage{algpseudocode}
\usepackage[mathlines]{lineno}
\usepackage{lipsum}

\usepackage{silence}

\newcommand{\ten}[1]{\mathcal{#1}}

\DeclareMathOperator*{\argmin}{argmin} 

\begin{document}
	\title{Determination of Latent Dimensionality in International Trade Flow}

\author{Duc P. Truong}
\address{Department of Mathematics, Southern Methodist University, Dallas, Texas, USA}
\author{Erik Skau}
\address{Computer, Computational, and Statistical Sciences Division, Los Alamos National Laboratory, Los Alamos, NM, USA}
\author{Vladimir I. Valtchinov}
\address{Department of Radiology and Center for Evidence based Imaging,
Brigham and Women's Hospital and Department of Biomedical Informatics, Harvard Medical School, Boston, MA, USA}
\author{Boian S. Alexandrov}
\address{Theoretical Division, Los Alamos National Laboratory, Los Alamos, NM, USA}
\ead{boian@lanl.gov}
\vspace{10pt}
\begin{indented}
\item[]January 2020
\end{indented}

\date{\today}

\begin{abstract}

Currently, high-dimensional data is ubiquitous in data science, which necessitates the development of techniques to decompose and interpret such multidimensional (aka tensor) datasets. Finding a low dimensional representation of the data, that is, its inherent structure, is one of the approaches that can serve to understand the dynamics of low dimensional latent features hidden in the data. Nonnegative RESCAL is one such technique, particularly well suited to analyze self-relational data, such as dynamic networks found in international trade flows. Nonnegative RESCAL computes a low dimensional tensor representation by finding the latent space containing multiple modalities. Estimating the dimensionality of this latent space is crucial for extracting meaningful latent features. Here, to determine the dimensionality of the latent space with nonnegative RESCAL, we propose a latent dimension determination method which is based on clustering of the solutions of multiple realizations of nonnegative RESCAL decompositions. We demonstrate the performance of our model selection method on synthetic data and then we apply our method to decompose a network of international trade flows data from \textit{International Monetary Fund} and validate the resulting features against empirical facts from economic literature.
  
\end{abstract}
\maketitle

\section{Introduction}
Dynamic networks are commonplace in many fields, such as economics, neuroscience, biology, recommender systems, data mining, and others. In these networks, different entities are represented by nodes, and their interactions over time are tracked through their edges. Dynamics networks are interrelated with the statistical relational models \cite{sharan2007exploiting} presented in artificial intelligence \cite{raedt2016statistical}. Statistical relational models often are characterized as: graphical models \cite{getoor2011learning}, latent class models \cite{magidson2004latent} and tensor factorization models \cite{nickel2011three}. Interestingly, there is a natural interconnection between the concept of latent models, graphical models and tensor factorization \cite{ishteva2015tensors,robeva2019duality}. The relational datasets are in the form of graphs, with nodes and edges representing, respectively, the various entities and their
relationships \cite{koller2007introduction}. The relational data is a graph-structured knowledge data that contain information for the 
relationships between some entities. 

\begin{figure}[h]
    \centering
    \includegraphics[width=1.0\textwidth]{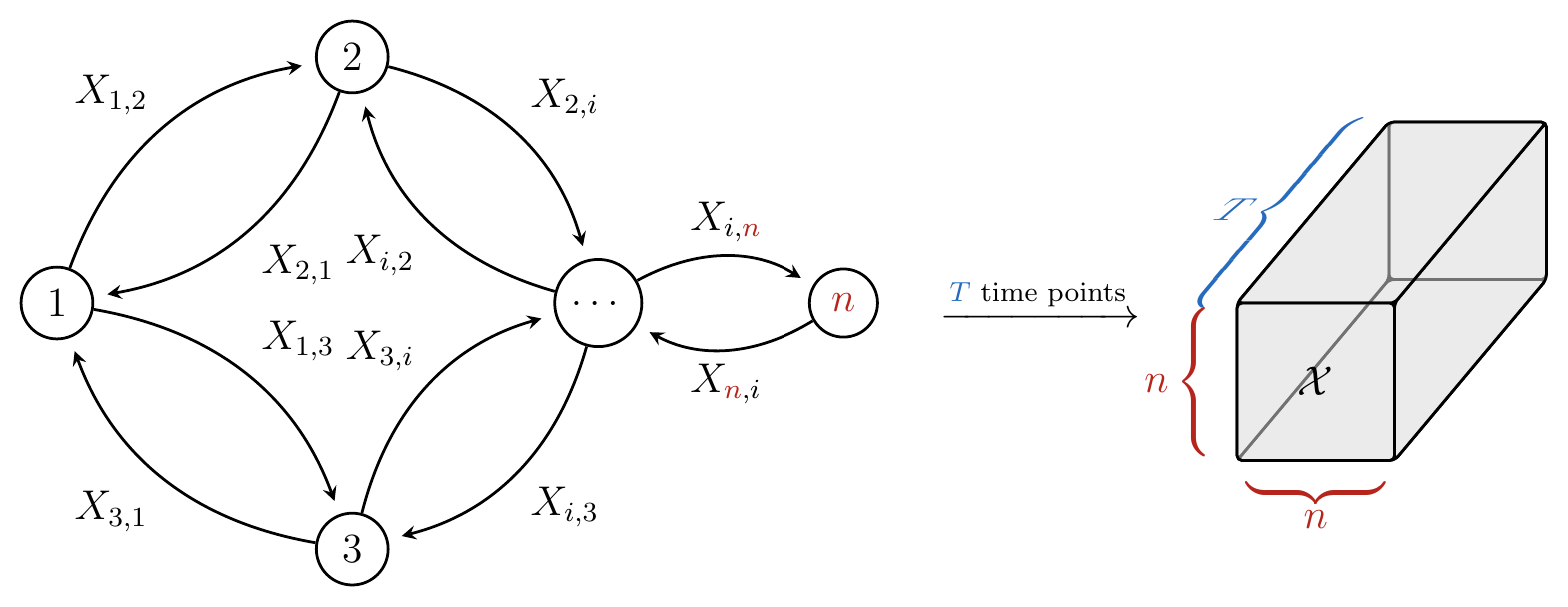}
    \caption{Dynamic network representing the interrelation between $n$ nodes, and the corresponding tensor.}
    \label{fig:dynamic_network}
\end{figure}

Tensor factorization methods have been proposed long ago for analyzing relational datasets and dynamic networks as the dynamics can be fully represented as three dimensional tensor with the first two axes indexing the nodes, and the third axis indexing time, Figure \ref{fig:dynamic_network}. It was shown that the RESCAL tensor decomposition can reveal notable interactions in dynamic asymmetric pairwise relationship tensor \cite{nickel2011three}, especially, when restricting the factors to be nonnegative, which make the model parts based and highly interpretable \cite{lee1999learning}.

One of the fundamental challenges in any model analysis is the determination of model hyperparameters \cite{mackay1994bayesian}. For the tensor decomposition this means selection of the correct latent dimension in the data that is related with the estimation of the rank of the analyzed tensor \cite{anandkumar2014tensor}, which is an NP-hard problem \cite{haastad1990tensor}. The development of  various procedures for model selection is an active research topic that include heuristics such as the Automatic Relevance Determination(ARD) \cite{morup2009automatic} or generalized class of information criteria for tensors with specific properties \cite{shi2019determining}. Especially in economics extracting easy understandable latent variables impacts important questions about the hidden mechanisms, causes, propagating channels, and groups of any international trade and economic events\cite{eaton2016trade,bems2013great}.

RESCAL can be considered as a specific nonnegative Tucker-2 decomposition with two equal factors, because of the inherent symmetry of the data. Hence, in RESCAL we need to know the multi-rank rather the rank of the analyzed tensor. In the case of nonnegative decomposition and even in presence of deficiency of the factors we can apply NMF to the corresponding unfoldings of the tensor, to find the minimal multi rank \cite{alexandrov2019nonnegative}. 

To determine the latent dimensionality in NMF, here we utilize a recent model determination technique, called NMFk, \cite{alexandrov2013deciphering, alexandrov2014blind} that is scalable \cite{chennupatidistributed}, and has been used to decompose the biggest collection of human cancer genomes \cite{alexandrov2013signatures}. NMFk integrate the classical NMF with custom clustering and Silhouette statistics \cite{rousseeuw1987silhouettes} and works on the principle of stability of the extracted latent variables, from several NMF-minimizations combined with the accuracy of the minimization in order to estimate the optimal number of latent features.
Here, we report the application of NMFk to determine latent dimensions, to the nonnegative RESCAL tensor decomposition. We demonstrate the efficacy of our method on synthetic data and apply our model determination method to decompose an well-known international trade flow dataset and validate our results against established economic empirical findings.

\subsection{Preliminaries and Notation}
Throughout this work vectors are denoted with lowercase bold letters, $  {x}$, matrices are denoted with uppercase letters $ {X}$, and tensors are denoted with uppercase script letters, $\ten{X}$. Mode-1 multiplication between an order three tensor and a matrix is defined $(\ten{X}\times_1  {Y})_{i,j,k} = \sum_{l=1}  {Y}_{i,l}   \ten{X}_{l,j,k}$, and similarly for modes 2 and 3. The Frobenius norm of a matrix or tensor is the square root of the sum of the squares of the elements, or $|| {X}||_F = \sqrt{\sum_{i,j}  {X}_{i,j}}$, $||  {X}||_F = \sqrt{\sum_{i,j,k} \ten{X}_{i,j,k}}$. A single subscript on a matrix, or tensor indicates the slice of the object along the last index, so $ {X}_i$ indicates the $i^{th}$ column of the matrix, and $  {X}_i$ indicates the $i^{th}$ matrix along the third mode. Additionally, a superscript in parentheses is used to enumerate items in a set or ensemble, $\{X^{(1)}, X^{(2)}, \hdots, X^{(p)} \}$.

\subsection{Related Works}
Here we describe some relevant works that are working with a similar decomposition model. The first proposed model was the single domain Decomposition into Directional Components (DEDICOM) model \cite{harshman1982model}. DEDICOM is capable of analyzing asymmetric data in marketing research and it has both matrix and tensor versions without constraints on its factors:
\begin{align*}
  &\text{ two-way DEDICOM} & {X} =  {A}  {R}  {A}^\top & \text{ where }  {A} \in  \mathbb{R}^{n \times r},  {R} \in    \mathbb{R}^{r \times r}  \\
  &\text{ three-way DEDICOM} &  \ten{X}_k =  {A}  \ten{D}_k  {R}  \ten{D}_k   {A}^\top &\text{ for }k = 1,\dots,T
\end{align*}
In the three-way DEDICOM model, \(\ten{X}_k\) is the \(k^{th}\) frontal slice of the tensor \(\ten{X}\), and \( \ten{D}_k \in  \mathbb{R}^{r \times r}\) is a diagonal matrix.
These models describe a \textit{single domain} in the sense that they require the row space to be the same as the column space. Since there is no constraint on the factors, these decompositions can be estimated similarly as SVD.

Later, an alternating algorithm, Alternating Simultaneous Approximation Least Square and Newton (ASALSAN) was proposed to perform a three-way DEDICOM \cite{bader2006temporal} that can be applied to large and sparse data. Moreover, the nonnegative version of three-way DEDICOM was also introduced using a multiplicative update algorithm. The model was then applied to analyze email network data and export/import data. However, the latent dimension was selected without being justified, and the meaning of the extracted factors was not analyzed in details.

A relaxed version of three-way DEDICOM is the RESCAL model \cite{nickel2011three}. The model decomposes a three-way tensor \(\ten{X}\) as follows:
\begin{align*}
    \ten{X}_k = A  \ten{R}_k A^\top \text{ for }k = 1,\dots,m
\end{align*}
In \cite{nickel2011three}, the factors \(A\) and \( \ten{R}_k\) minimized the \textit{\(l_2\)-regularized} minimization problem, and can be estimated using a variation of ASALSAN:
\begin{align*}
  \min_{A, \ten{R}_k} = \dfrac{1}{2}\left(\sum_k || \ten{X}_k - A \ten{R}_k A^\top ||^2_F\right) + \dfrac{1}{2}\lambda \left( ||A||^2_F + \sum_k || \ten{R}_k ||^2_F \right)
\end{align*}
The model was then applied to different datasets, and the authors suggested to find latent dimension using cross-validation. Instead, they chose a rather large dimension (\(k=20\)), and then used k-means clustering method to cluster the matrix \(A\) into 6 groups. 

Following the initial work, Ref. \cite{krompass2013non} introduced updating schemes for different variants of the nonnegative RESCAL mode, including least-squares with \(l_2\) regularization, KL-divergence with \(l_1\) regularization, and other.

\section{Methods and Implementation}
\subsection{Nonnegative RESCAL}
The RESCAL model is a tensor decomposition that takes advantage of the inherent structure of relational properties such as those expected in a dynamic network. More specifically, RESCAL decomposes an order three tensor by finding a common low dimensional latent space for the first two modes such that $$ \ten{X} = \ten{R} \times_1 A \times_2 A$$ where $A$ is an $n \times r$ matrix containing the features, and $ \ten{R}$ is an $r \times r \times T$ tensor capturing the mixing relations between the features. In practice, a decomposition is always approximated by solving the optimization problem \begin{align}
    \argmin_{ {A},  \ten{R}} ||   \ten{X} -   \ten{R} \times_1  {A} \times_2  {A}||_F^2\;.
  \end{align}  When applied to a dynamic networks, RESCAL is interpretable in the way that each column of \(A\) represents a  group of objects or nodes, and \(\ten{R}_t\)  represents the relations among these groups at time \(t\). An equivalent problem statement demonstrates that RESCAL simultaneously decomposes each slice of an order three tensor with a rank $r$ factorization, \begin{align}
  \argmin_{ {A},  {R}_t} \sum_t ||   \ten{X}_t -  {A}   \ten{R}_t  {A}^{\top} ||_F^2\;.
  \end{align} 
  With this interpretation RESCAL attempts to find a common set of features that can be simultaneously used to represent both the row space and the column space of the matrices $ \ten{X}_t$. To remove a scaling ambiguity, RESCAL also can constrain the columns of $A$ to be unit norm,  $|| A_i || = 1$ for $1 \leq i \leq r$.

Quite often with nonnegative data, a nonnegative decomposition provides more insightful features with parts based decomposition \cite{lee1999learning}. Nonnegative RESCAL provides the same advantage by decomposing the data with nonnegative features, and nonnegative mixing matrices with the optimization
\begin{equation}
  \label{eqn:minproblem}
  \begin{aligned}
  \argmin_{ {A},  \ten{R}_t} & \quad \sum_t ||  \ten{X}_t -  {A}   \ten{R}_t  {A}^{\top}
  ||_F^2\\
  \text{subject to} & \quad \begin{cases} \sum_j A_{ij} = 1, \text{ for } 1 \leq j \leq r \\ A \geq 0\\  \ten{R} \geq 0 \end{cases}\;\end{aligned} \end{equation}.
\begin{figure}[h]
  \centering
  \includegraphics[width = 0.8 \textwidth]{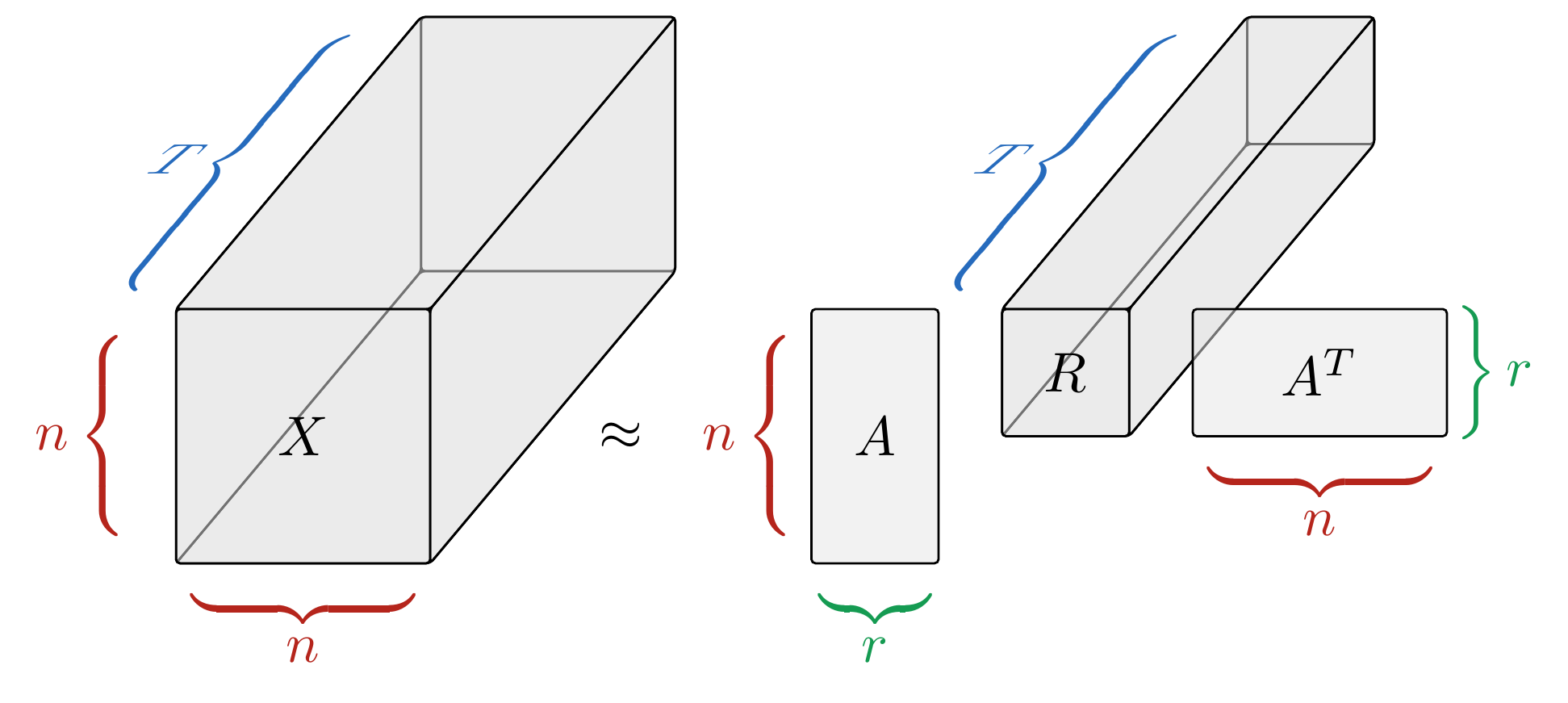}
  \caption{RESCAL Model - Columns of \(A\) specify groups of objects, and tensor \(\ten{R}\) captures group interactions through time.}
  \label{fig:symtucker}
\end{figure}

\subsection{Multiplicative Update Algorithm } 
To solve the nonnegative constraint minimization problem in equation \ref{eqn:minproblem}, we use the multiplicative update scheme similar to the one used for DEDICOM model \cite{bader2006temporal}, and for nonnegative RESCAL \cite{krompass2013non}. Starting from nonnegative random initialization of \(A\) and  \(R\), the following update steps are performed until the relative error converges to a predefined tolerance:
\begin{equation}
  \begin{gathered}
    Vec(\ten{R}_t)_i = \dfrac{Vec(\ten{R}_t)_i \left[Vec( {A}^\top \ten{X}_t {A} )\right]_i}{\left[ {A}^\top  {A} \otimes  {A}^\top  {A} Vec( \ten{R})\right]_i + \epsilon} \text{ \ \ for } t = 1,..,T \\
     {A}_{ij} =  {A}_{ij}\dfrac{\left[\sum_{t=1}^T  \ten{X}_t {A} \ten{R}_t^\top +  \ten{X}_t^\top  {A}   \ten{R}_t\right]_{ij} }{\left[ {A}\left( \ten{R}_t {A}^\top  {A} \ten{R}_t^\top +  \ten{R}_t^\top  {A}^\top  {A}   \ten{R}_t \right)\right]_{ij} + \epsilon}
  \end{gathered}
\end{equation}
where \(Vec(\cdot)\) is the vectorize operator.
After the multiplicate update scheme converges, \( {A}\) and \( \ten{R}\) are appropriately scaled such that columns of \( {A}\) have a sum equal to one. Notice that the decomposition can be scaled without affecting the reconstruction error.

\subsection{Model Selection} \label{sec:model_selection}
\begin{figure}[h]
  \centering
  \includegraphics[width = 1.0 \textwidth]{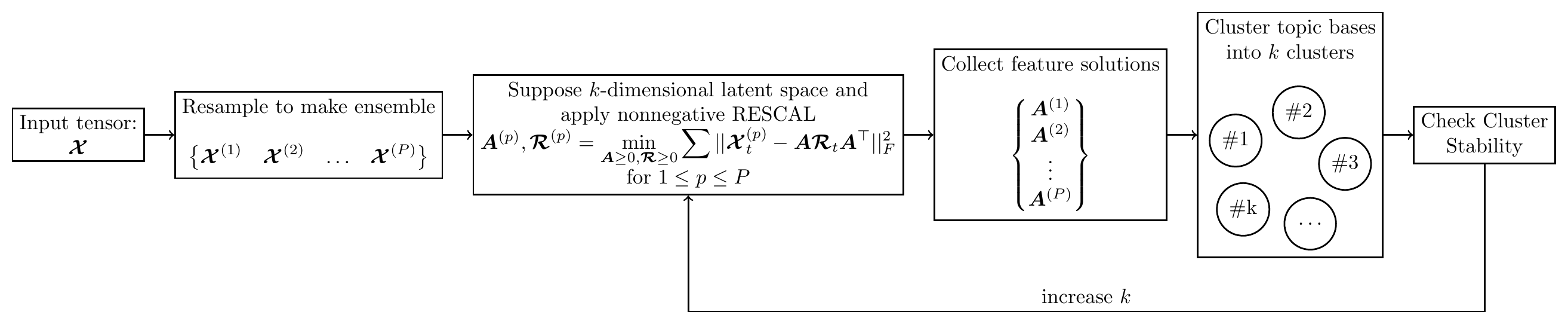}
  \caption{Diagram of steps to determine latent dimension for nonnegative RESCAL.}
  \label{fig:rescal-k}
\end{figure}

To select an appropriate latent dimension, we adapt the clustering procedure that has found success in NMF \cite{alexandrov2013deciphering}. For each explored latent dimension, $k$, our procedure applies three steps: (i) bootstrapping by resampling the data, (ii) decomposing the bootstrapped data, and then (iii) analyzing the cluster stability of the solutions. Figure \ref{fig:rescal-k} shows a diagram of the model selection scheme.
To resample the data, we construct an ensemble of tensors $\{ \ten{X}^{(1)},  \ten{X}^{(2)}, \hdots,   \ten{X}^{(P)}\}$ sampled from $ \ten{X}^{(p)}_{i,j,k} \sim  {U}(1-\epsilon, 1+\epsilon) *   \ten{X}_{i,j,k}$ for $1 \leq p \leq P$ where $ {U}(a,b)$ is a uniform distribution between $a$ and $b$. Each resampling introduces some variability in the data which mitigates the possibility of overfitting.

We use custom clustering algorithm designed to exploit the stability of the NMF's solutions corresponding to the resampled data. The custom property of the clustering is that each cluster should contain precisely one feature vector from each $A^{(p)}$. Our algorithm is based on k-means but iterates over each $A^{(p)}$ to assign each vector to an appropriate centroid. The assignment is done by a greedy algorithm applied to the cosine similarity between the columns of $A^{(p)}$ and the current centroids.

To analyze the quality of the clustering, we use silhouette statistics \cite{rousseeuw1987silhouettes}. The silhouette of a single point has a value between -1 and 1 relating how close it is to other points in the same cluster and how far is this point to the closest of the other clusters. A high silhouette score indicates that the clusters are compact and well separated, while a low silhouette score indicates that the clusters are not well separated. We use both the mean silhouette score, as well as the minimum silhouette score of a group as cluster quality metrics.

The selection of the correct latent dimension is accomplished by considering both the silhouette statistics, which measures the stability of the solutions, and the relative error. We expect that with the correct latent dimension the solutions will cluster well with a small relative error. With a too small latent space, the relative error will be too large, and with too large latent space there will be latent features representing the noise in the system that are not stable and hence will not cluster well, resulting in a low silhouette score. We consider the correct latent dimension to be the largest dimension that generates low relative errors and high silhouette scores.

\section{Experiments}
\subsection{Synthetic Data}
\label{sec:syntheticdata}

Here we test the performance of our protocol for model selection in determining the ground truth latent dimension of synthetic datasets with different levels of noise. First, a synthetic data tensor \( \ten{X}\) with dimensions \(n \times n \times T\) and the latent dimension \(k\) is generated as follows:
\begin{itemize}
  \item Elements of matrix \(A_{n \times k}\) are randomly sampled from \(  {U}[0,10)\). The matrix \(A\) is only selected if \(rank(A) = k\). Otherwise, it is regenerated.
  \item Elements of matrix \(A\) is sparsified by setting elements smaller than \(Thresh_{A}\) to zeros.
  \item The tensor \( \ten{R}\) is also generated from \(  {U}[0,10)\).
  \item Tensor \( \ten{R}\) is also sparsified by setting elements smaller than \(Thresh_{R}\) to zeros.
  \item \( \ten{X} = \ten{R} \times_1 A \times_2 A + {\epsilon}\) where elements of $\epsilon$ are sampled from \( {U}[0,NoiseFactor)\).
  \item The noise level of the data tensor can be adjusted by the value of \(NoiseFactor\).
  \item The noise level is computed as \(\dfrac{|| \ten{X} - A \ten{R}A^\top||_F}{||A \ten{R}A^\top||_F} \).
\end{itemize}
The latent dimension determination procedure described in \ref{sec:model_selection} is then applied to the simulated data. More specifically, the sample perturbation is $0.03$ and the number of iterations \(P = 50\).
\begin{figure}[h]
  \centering
  \includegraphics[width=1.0\textwidth]{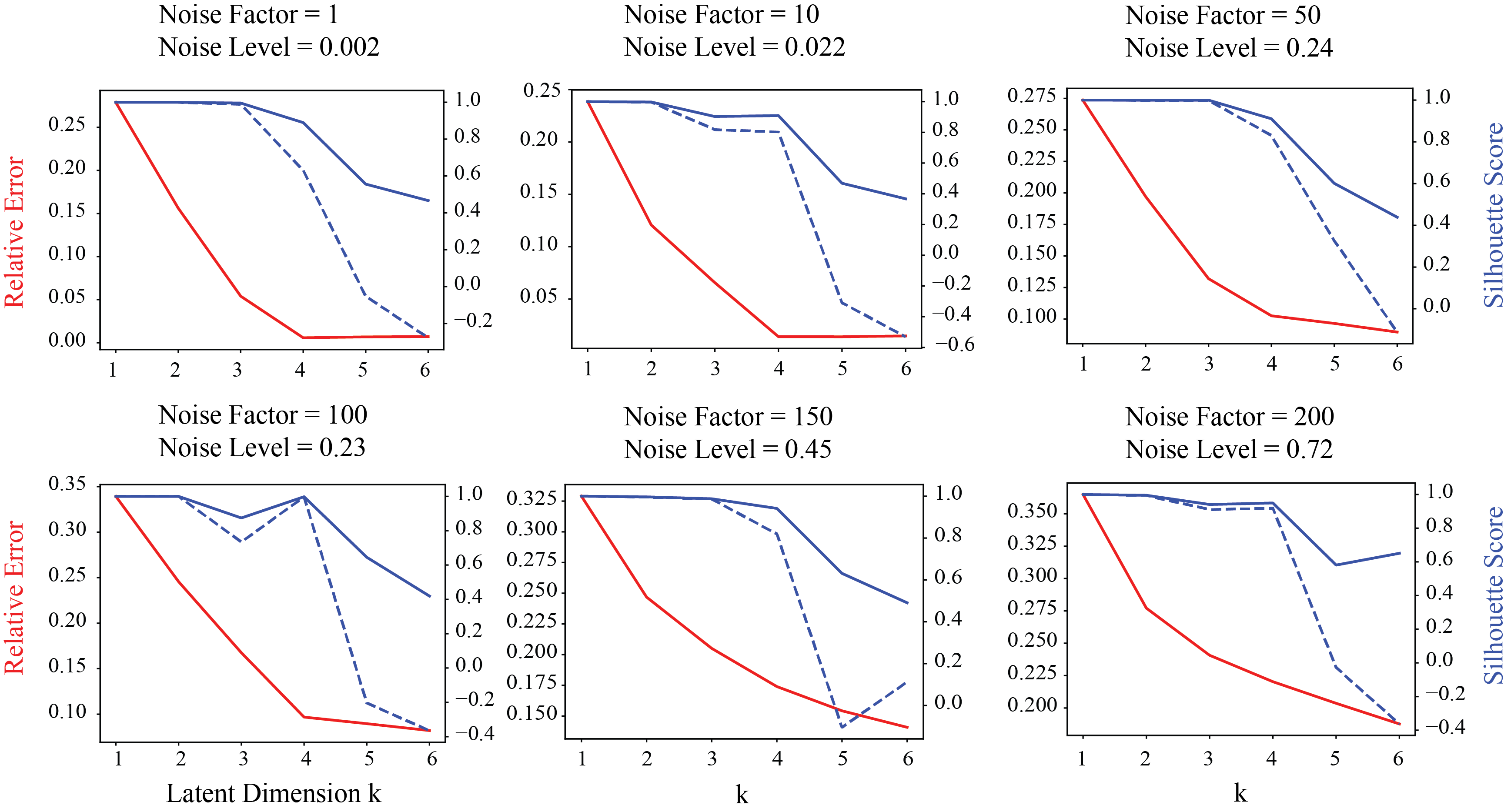}
  \caption{The indicator from silhouette score curves to determine the true latent dimension is consistent across different noise levels. \textit{Tensor dimension \(10 \times 10 \times 100\), \(k_{true} = 4\), \textcolor{red}{(Red)}: Relative reconstruction error, \textcolor{blue}{(Blue Solid)}: Average Silhouette Score, \textcolor{blue}{(Blue Dash)}: Minimum Silhouette Score}  }
  \label{fig:sparse38}
\end{figure}

In these scenarios, for a range of \(NoiseFactor \in \{1,10,50,100,150,200\}\), a tensor of dimension \(10 \times 10 \times 100\) is simulated with the ground truth dimension \(k_{true} = 4\). The plots of relative reconstruction error, minimum and average silhouette scores are shown in Figure \ref{fig:sparse38}. As the noise factor increases, the noise level also increases and gradually distorts the L-shape of the reconstruction error curves. However, the silhouette score curves are consistent up to \(k=4\), and after that, the silhouette score curves start diverging, demonstrating that  the clusters are no longer compact and well separated.

\subsection{Economic Application: Decompose the International Trade Flows}

International trade has been shown to generate mutual benefit between countries by allowing them to focus on specialization and exchange their produced goods and services. Unsurprisingly, it has increasingly contributed to the Gross Domestic Product (GDP). In fact, according to the World Bank, in 2017, the world GDP is about \$80 trillion. Of this total, about 29 percent was traded across countries: international trade flows in goods and services is about \$23 trillion. It has been shown that the economic growth of a country is strongly associated with its role in the world trading network. Therefore, understanding the trade pattern is an essential step before we can discuss the effects of international trade, or even suggest policy changes. 

Here we show that by applying the nonnegative RESCAL model, with our latent dimension determination method, we can decompose the international trade network into different groups of countries whose exports and imports are tightly linked together and their interactions over time are captured in the resulting interacting tensor. More interestingly, the groups' activities are also meaningful in the sense that they are consistent with stylized economic facts.

\subsubsection*{International trade flows data}
We obtained the international trade flows data from \textit{Direction of Trade Statistics, IMF} \cite{marini2018new}. Specifically, the data contains monthly export amounts in U.S. dollars between 23 countries from January 1981 to December 2015 (420 months). Thus the data tensor has 23 by 23 by 420 dimensions, in which each entry, $\ten{X}_{ijk}$, represents how much country \(i\) exports to countries \(j\) in month \(k\). For clarity, the list of countries includes Australia, Canada, China Mainland, Denmark, Finland, France, Germany, Hong Kong, Indonesia, Ireland, Italy, Japan, Korea, Malaysia, Mexico, Netherlands, New Zealand, Singapore, Spain, Sweden, Thailand, United Kingdom, United States.

This dataset has previously been analyzed with variations of the RESCAL model. In \cite{chen2017factor}, Chen \textit{et al.} also apply the RESCAL model, though not nonnegative, and they do not have a procedure to determine the correct latent dimension, they simply chose a latent dimension of three for illustrative purposes of the model.

\subsubsection*{Determine the latent dimension \& optimal factors}
\begin{figure}[h]
  \centering
  \includegraphics[width = 0.8\textwidth]{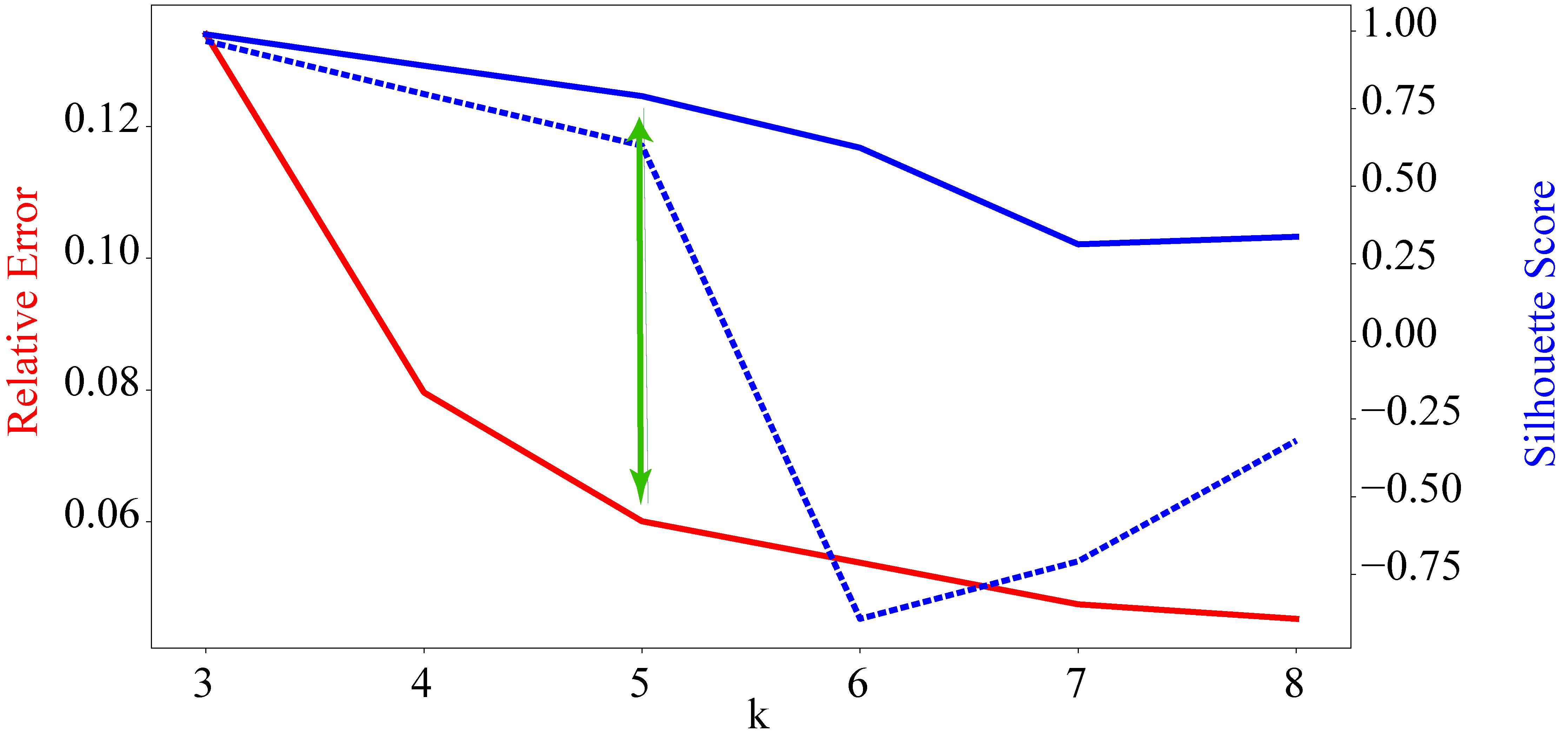}
  \caption{\textit{(Red)} Relative reconstruction error in Frobenius norm. \textit{(Solid blue)} Average Silhouette Score. \textit{(Dashed Line)} Minimum group silhouette score. The largest gap between the relative reconstruction error curve and the silhouette statistics occurs at \(k=5\), which indicates that 5 is the optimal number of groups to explain the data.}
  \label{fig:errorscore}
\end{figure}

To determine the latent dimension for the model, we resampled as described in Section \ref{sec:model_selection} with the uniform distribution $ {U}(0.9, 1.1)$ so that each value in our ensemble has the measured value $\pm 10\%$ error. We resampled from this distribution 50 times to construct our ensemble and calculated the silhouette statistic for a range of dimensions \(k \in \{3,\dots,8\}\). Figure \ref{fig:errorscore} shows the relative reconstruction errors, the average and minimum silhouette statistics leading us to conclude that that the true latent dimension is 5.

To determine the optimal factors for the analysis, we first ran 100 iterations from random initialization on the original data with the selected dimension \(k=5\), and select the decomposition, which provided the lowest reconstruction error. For both purposes, the stopping criterion is the relative convergence rate \(= 1e-8\).

\subsubsection*{Interpretation of the decomposition}
\begin{figure}[]
  \centering
  \includegraphics[width = 0.7\textwidth]{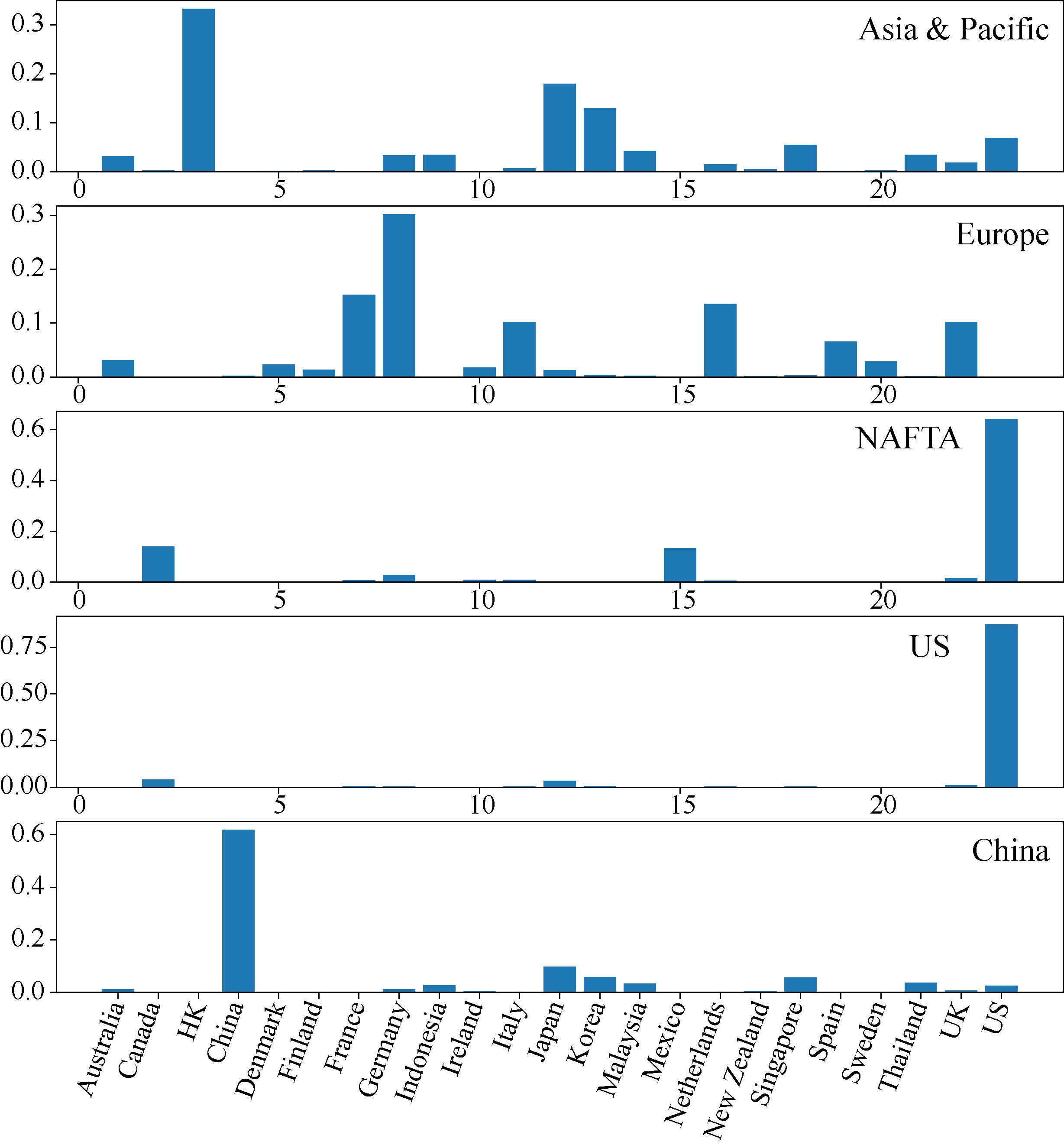}
  \caption{Normalized columns of A, latent factors, show the relative contribution of the countries in each group. Based on the major contributions, five latent factors approximately represent five actual economic regions : Asia and Pacific (without China), Europe, NAFTA (Canada, Mexico, and United States), United States, and China.}
  \label{fig:groups}
\end{figure}

Here we show the interpretation and analysis of the optimal decomposition. First, since the columns of \( {A}\) are normalized to have a sum equal to one, the entry $ {A}_{ij}$ can be interpreted as how much the country $i$ contributes into group $j$. Figure \ref{fig:groups} shows that the latent factors, which here are the country contribution profile in each group, approximately correspond to five economic regions: Asia and Pacific (without China), Europe, NAFTA (Canada, Mexico, and the U.S.), U.S., and China. These Geo-economic regions are essential participants in the international trade, verifying that the model selection procedure extracts meaningful latent factors.

Next, we analyze the interacting tensor \( \ten{R}\) to determine its economic meanings. We first look at the aggregate export level for each economic group over time by summing across the rows of each \( \ten{R}_t\) without the diagonal elements. By doing this, we exclude the contribution of the groups themselves to their aggregate export activities. We then check how well these approximated export activities agree with four global and local economic recession periods, which are identified by the National Bureau of Economic Research (NBER). 

\begin{figure}[h]
  \centering
  \includegraphics[width=1.0 \textwidth]{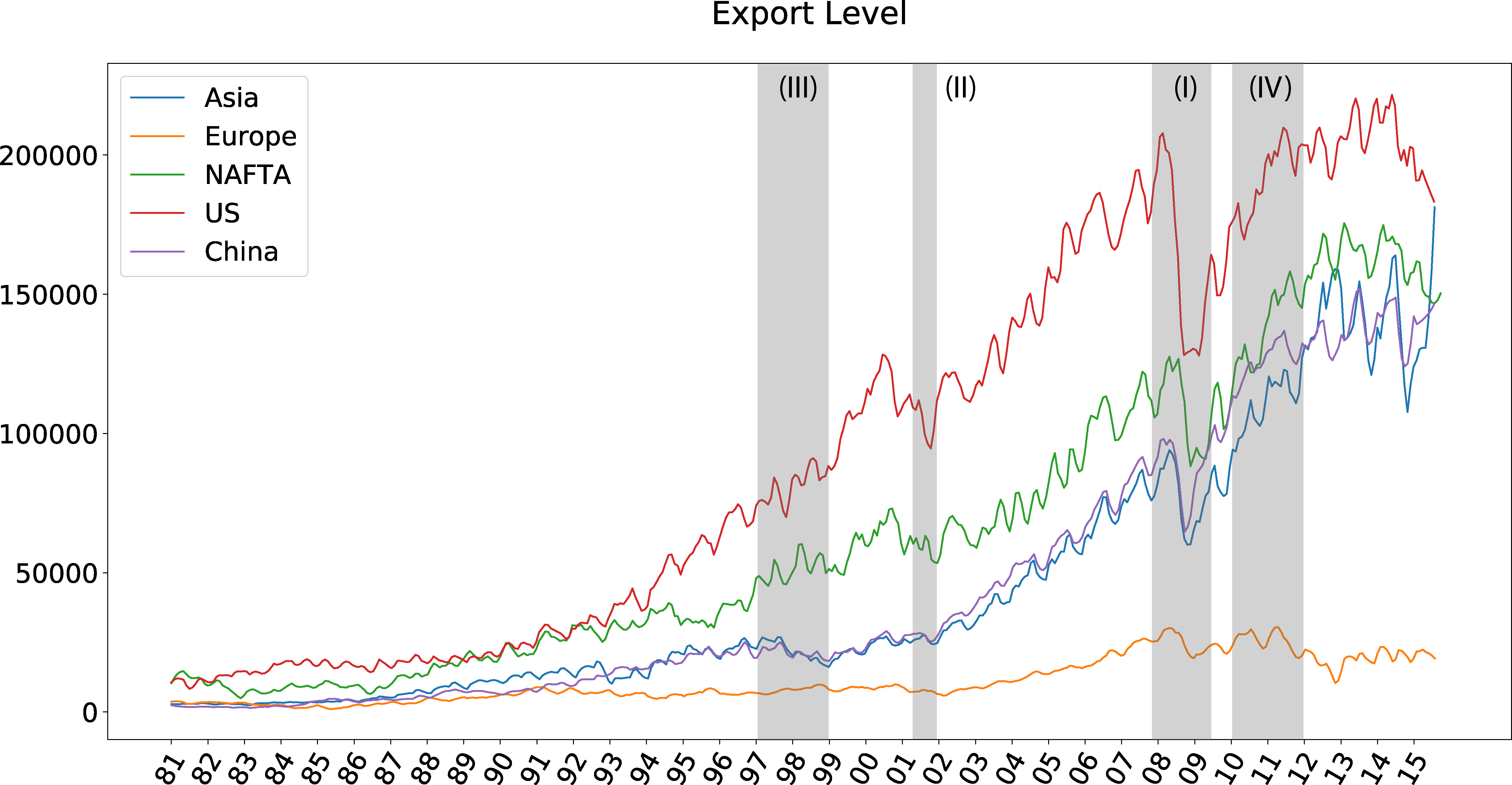}
  \caption{Export activities of each group. Grey areas are periods of economic recessions. (I) Great Recession: This is a global recession that affected all groups. (II) Early 2000s recession, which only affected developed countries. (III) Asian Currency Crisis 1997-1999, which only affected Asia. (IV) European Debt Crisis 2010-2012 which affect European countries.}
  \label{fig:export level}
\end{figure}

As shown in Fig \ref{fig:export level}, the approximated activities well match with the international trading trends in all considered periods. For example, during the \textit{Great Recession} (12/2007-06/2009), in which international trade was dramatically decreasing, the activities of all groups dropped substantially. Secondly, during \textit{early 2000s Recession}, which was partially caused by the dotcom bubble and September 11, the activities represent the fact that the trading trend of all groups, except Asia, were dropping. Thirdly, during the \textit{Asia Financial Crisis (1997-199)}, only the activities of the group Asia is going down, representing the fact that this Recession mostly affected Asian countries. Lastly, during the \textit{European Debt Crisis}, which peaked in 2010-2012, while the economy of other regions was recovering from the Great Recession, only European countries struggled with their high government debt. This styled fact is replicated in the approximated activity of the Europe-group. Overall, the interacting tensor captured the connection between the export and economic health of different economic regions.

Finally, we analyze the interaction tensor \(R \in \mathbb{R}_{5 \times 5 \times 420}\) by summing the tensor \(R\) over time, which gives us the matrix \(S \in \mathbb{R}_{5 \times 5}\) describing how strongly these groups interact. \(Rs\) is then normalized by its maximum value for better visualization. 
\[\bar{S} = S/max(S) \text{ where }S_{i,j} = \sum_{t=1}^{420} R_{i,j,t}\]

\begin{figure}[h]
  \centering
  \includegraphics[width= 1.0 \textwidth]{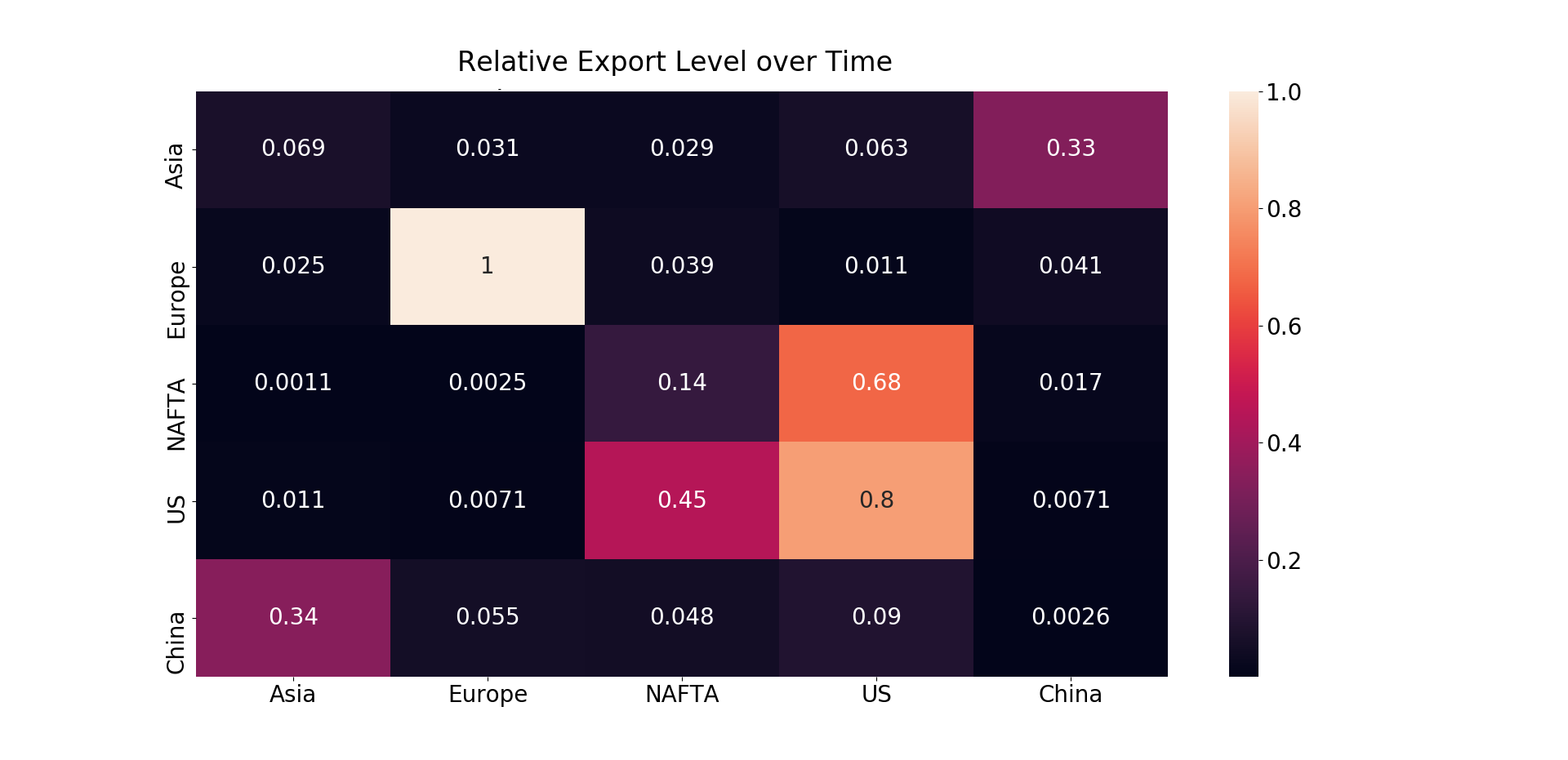}
  \caption{Normalized Export Level between groups}
  \label{fig:heatmap-export}
\end{figure}

We makes three observations from Figure \ref{fig:heatmap-export}. First, the Europian group has the strongest interaction with itself, which makes sense since this is a group of advanced economies, and they have formed the European Union since 1995. Second, the strong two-way connection between NAFTA and the United States and between China and Asia, are also matched with statistics. Third, by comparing the row and column for China and United States, we can see that China is a trade surplus (the amount of exports is greater than the number of imports), and the United States is a trade deficit. Overall, the interacting tensor did extract meaningful information from the data in the sense that they agree with international economics empirical facts, indicating that our model selection procedure can capture meaningful activities from the data. 

\section{Discussion}
In this paper, we introduce a model selection protocol for determination of the dominant latent dimension of the nonnegative RESCAL model. This method, which is based on nonnegativity assumptions on both factors \(A\) and \(  \ten{R}\), evaluates the stability of the latent factors \(A\), or equivalently the quality of clusters generated by factorization of a set of different realizations of the input data. The method then selects the highest dimension at which the stability is still high. Our method performs well on sparse synthetic data with different noise levels. Moreover, when applied to a real dataset, the international trade flows data from \textit{IMF}, the model was able to decompose considered countries into meaningfull geo-economics regions; with interacting activities matching the trading characteristics of each region and consistent with what has been observed about different economic recessions.

It would be of a particular interest to apply the method presented here to two common modeling challenges in business analytics and quantitative marketing: forecasting, and customer marketing segmentation. For example, traditional forecasting techniques rely almost exclusively on the time-series properties of the learning data set (usually called statistical forecasting methods). Another set of techniques that have been developed introduces additional (external) variables, and a regression-like model was fitted with the additional requirement that the residuals are an ARIMA-distributed process; they are referred to as machine learning (ML) based forecasting methods \cite{ahmed2010}. However, it is unclear which set of techniques is the better one and would universally work for different types of forecasts: customers forecasting vs. revenue forecasting vs. i.e., inventory forecasting. Two recent papers have carried out an ad-hoc comparison of statistical vs. machine learning models and have arrived at precisely opposite conclusions using similar testing methodologies and goodness of fit metrics (\cite{ahmed2010} and \cite{markidakis2018}). Both statistical and machine learning techniques use the time-series nature of the data in the forecasting in a specific manner, whereas the method presented here treats the time dimension like all other dimensions of the multidimensional data set. Further, it is of interest to compare the performance of our method for forecasting of customers and revenue and contrast it against the statistical and the ML methodologies. Remarkably, Markidakis et al. (\cite{markidakis2018}) have also compared the forecasting performance of several Deep Learning algorithms and have found it to be stacking unfavorably against the statistical models -- it would be instructive also to see how the Nonnegative RESCAL algorithm fares against deep learning models. 

Further, the customer segmentation is a fundamental modeling exercise in quantitative and precision marketing. Customer segmentation has almost exclusively been done in two distinct, loosely connected steps: A) segmenting customers using static data (no time-resolved data, typically using clustering techniques), B) consider transitions to (slightly) different segmentation states, to simulate time-resolved behavior. We intend to apply the Nonnegative RESCAL methodology to this problem as the time dimension is treated in much more natural fashion here. 

\section*{Acknowledgment}

The work performed in Los Alamos National Laboratory is supported by the U.S. Department of Energy National Nuclear Security
Administration under Contract No. DE-AC52-06NA25396 and LANL laboratory directed research and development (LDRD) grant 20190020DR. Computations were supported by LANL Institutional Computing Program.

\newpage
\section{Bibliography}
\bibliography{references}
\bibliographystyle{iopart-num}

\end{document}